\documentclass[conference]{IEEEtran}
\IEEEoverridecommandlockouts
\usepackage{cite}
\usepackage[colorlinks,linkcolor={blue},citecolor={red},urlcolor={red}]{hyperref}
\usepackage[utf8]{inputenc}
\hypersetup{hidelinks}
\usepackage{amsmath,amssymb,amsfonts}
\usepackage{algorithmic}
\usepackage{graphicx}
\usepackage{textcomp}
\usepackage{booktabs}
\usepackage{makecell}
\usepackage{lipsum}  
\usepackage{tabularx}
\DeclareUnicodeCharacter{202C}{\,}
\usepackage{float}
\usepackage{subcaption}
\usepackage[printonlyused, withpage]{acronym}
\usepackage{xcolor}
\usepackage[colorlinks]{hyperref}
\def\BibTeX{{\rm B\kern-.05em{\sc i\kern-.025em b}\kern-.08em
    T\kern-.1667em\lower.7ex\hbox{E}\kern-.125emX}}

\begin{document}
\title{ Physical Annotation for Automated Optical Inspection: A Concept for In-Situ, Pointer-Based Training Data Generation}

\author{\IEEEauthorblockN{1\textsuperscript{st} Oliver Krumpek}
\IEEEauthorblockA{\textit{Dept. Machine Vision} \\
\textit{Fraunhofer IPK}\\
Pascalstr. 8-9, 10587 Berlin, Germany \\
oliver.krumpek@ipk.fraunhofer.de}
\and
\IEEEauthorblockN{2\textsuperscript{nd} Oliver Heimann}
\IEEEauthorblockA{\textit{Head of Dept. Machine Vision} \\
\textit{Fraunhofer IPK}\\
Pascalstr. 8-9, 10587 Berlin, Germany}
\and
\IEEEauthorblockN{3\textsuperscript{rd} Prof. Jörg Krüger}
\IEEEauthorblockA{\textit{Chair Industrial Automation Technology}} \
\textit{Technical University Berlin}\\
Pascalstr. 8-9, 10587 Berlin, Germany }

\maketitle

\begin{abstract}
This paper introduces a novel physical annotation system that is designed to generate training data for automated optical inspection. The system uses pointer-based, in-situ interaction to transfer the valuable expertise of trained inspection personnel directly into a machine learning training pipeline. Unlike conventional screen-based annotation methods, our system allows annotation directly on the physical object, providing a more intuitive and efficient way to label data. The core technology uses calibrated, tracked pointers to accurately record user input and convert these spatial interactions into standardised annotation formats compatible with open-source software. A simple projector-based interface also projects visual guidance onto the object to assist users during the annotation process, ensuring greater accuracy and consistency. The proposed concept bridges the gap between human expertise and automated data generation. It enables non-IT experts to contribute to the ML training pipeline. Preliminary evaluation results confirm the feasibility of capturing detailed annotation trajectories and demonstrate that integration with CVAT streamlines the workflow for subsequent ML tasks. This paper details the system architecture, calibration procedures, and interface design, and discusses the concept's potential contribution to future ML-based automated optical inspection.

\end{abstract}

\section{Introduction}

\begin{figure}[htbp]
\centering
\includegraphics[width=0.44\textwidth]{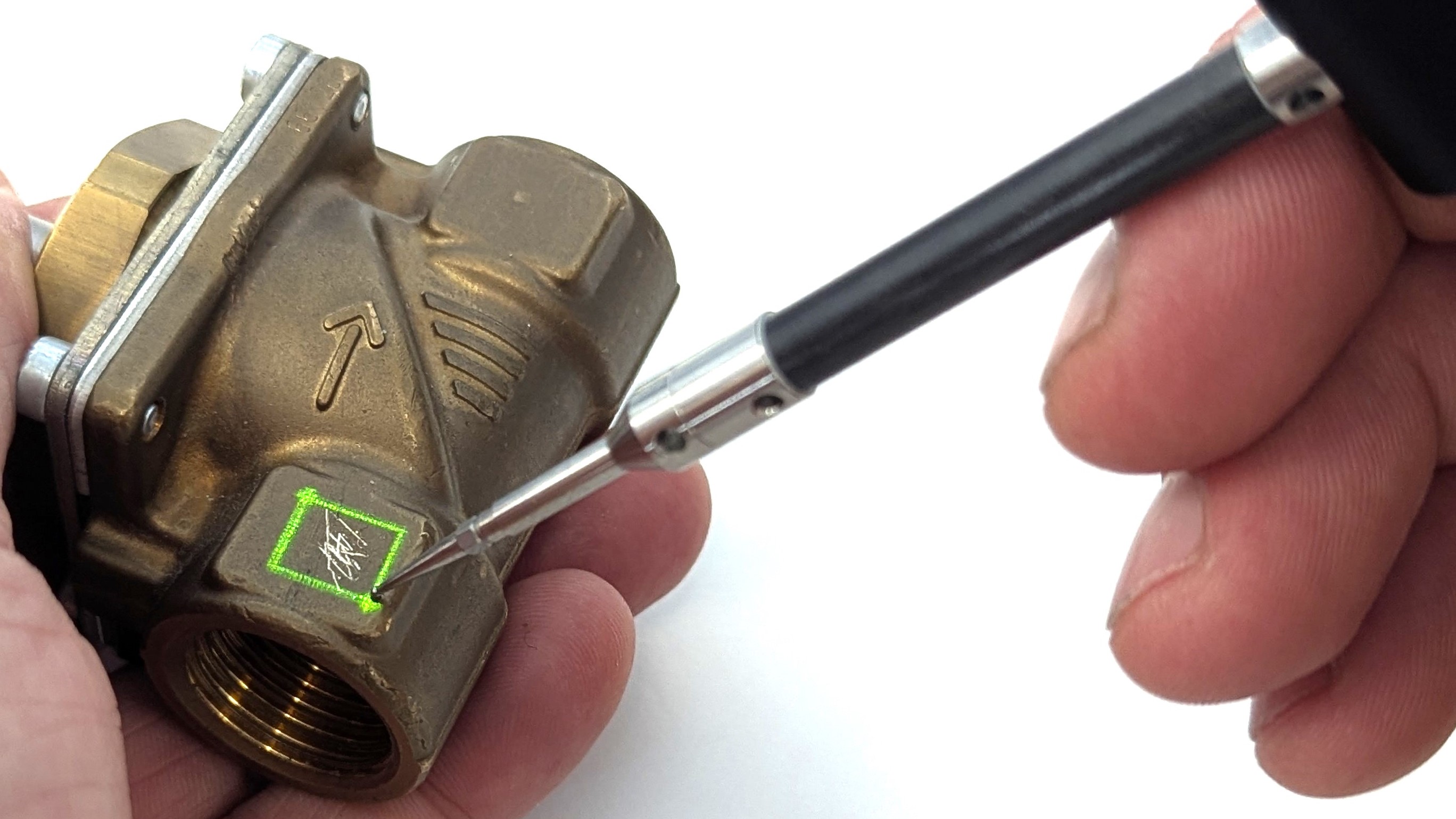}
\caption{Pointer based labeling of a visible defect on the surface of a mechanical part in a manual inspection process using the bounding box feature. }
\label{fig:hardware_setup}
\end{figure}

\begin{figure*}[htbp]
  \includegraphics[width=1\textwidth]{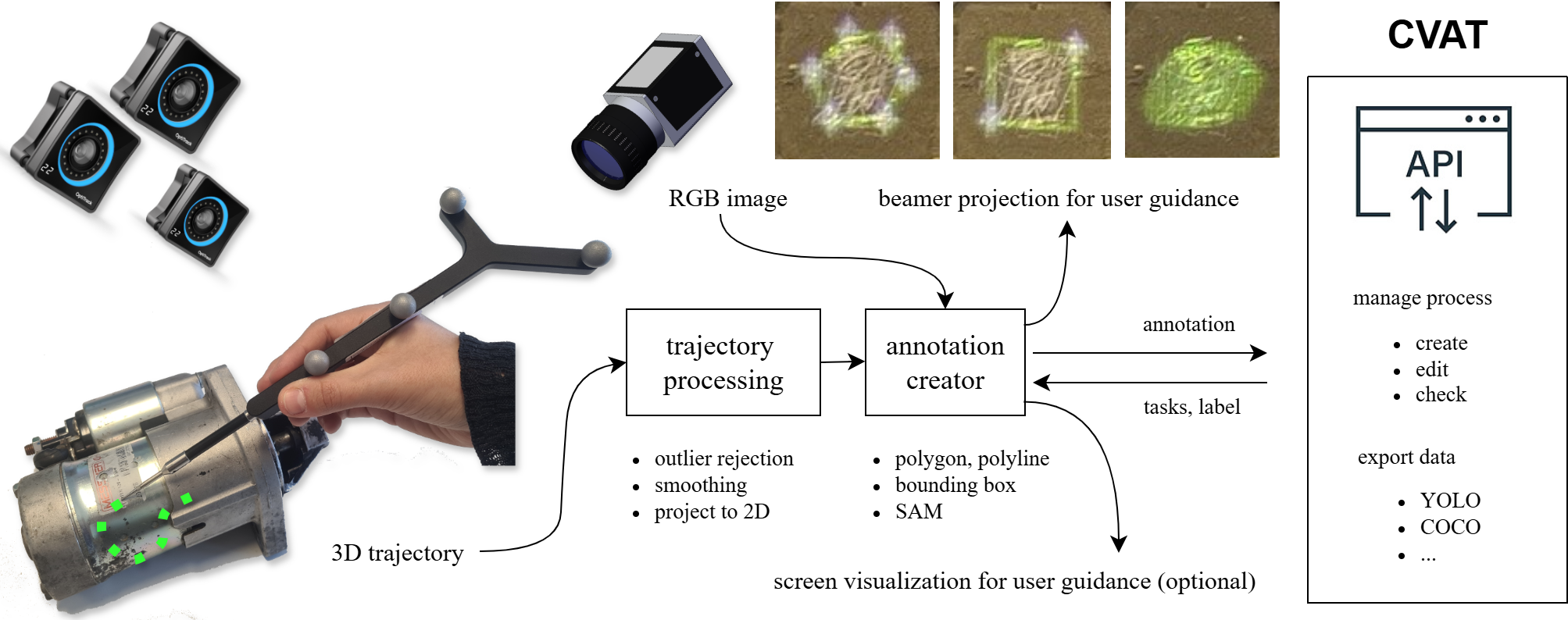}
  \caption{Illustration of the system's components, along with the work- and dataflow involved in generating training data.}
\end{figure*}

Recent advances in AI models have enabled a multitude of novel optical inspection tasks for industrial use. When properly trained, AI-driven systems can facilitate manual processes or, depending on the complexity and requirements, fully automate inspection tasks. However, the performance of these systems depends heavily on the quality and accuracy of the training data. Collecting such data in real-world scenarios involving human inspectors remains challenging. In many projects, the lack of manually labelled training data hinders the implementation of robust automated inspection systems. In industrial quality inspection, the number of defective parts can be relatively small compared to the total number of objects inspected. This is particularly true in safety-critical quality control, where a 100\% detection rate is required for parts to meet high production standards. Additionally, the process of digitising and annotating defects is typically neither straightforward nor well integrated. It often requires the use of external digitisation devices and supplementary software. Furthermore, inspection processes are often well established and have been in place for many years. Modifying them typically necessitates financial investment and high-level management approval. This aligns with current research, which identifies insufficient IT skills and data availability as the main challenges in implementing AI in manufacturing \cite{Heimberger.2024}. 
To address these challenges, we propose a physical, in-situ annotation system that captures expert knowledge directly within the inspection process. Leveraging intuitive, pointer-based interactions, our approach allows inspectors to annotate classes and segments directly on the object's surface. The underlying concept and objective of the development is to create an inspection system that facilitates the continuous and mutual exchange of expert knowledge between human inspectors and AI expert systems. The system aims to improve the quality and robustness of inspection tasks in this way. It also increases acceptance of AI in the process by making the data generation process more transparent and integrating it into existing workflows. This paper details the system architecture, calibration procedures and interface design, demonstrating how our approach bridges the gap between expert insight and the acquisition of effective ML training data in manufacturing environments.

\section{Contributions} \label{sec:contributions}

We present a novel in-situ approach to generating training data, which is specifically designed to address real-world problems in manual and assisted optical inspection tasks. Our approach enables non-IT experts to generate data through intuitive, pointer-based interaction — simply by pointing at the target object, without the need for screen-based annotation tools. The system is designed to save time and prevent valuable training samples from being overlooked by integrating the annotation process directly into the inspection workflow.


\section{Related Work} \label{sec:relatedwork}

Human-in-the-loop annotation methods, which are the focus of this article, emphasise semi-automatic and directly supervised approaches. While incorporating human expertise increases the time and effort required, it ultimately improves the quality and accuracy of the training data. These methods rely on screen-based interfaces such as CVAT, LabelImg, Supervisely and the VGG Image Annotator, all of which support a wide range of annotation techniques for 2D, 3D and multimodal data. For a more intuitive annotation process the paper LookHere \cite{Zhou.2022} presented a gesture-aware approach using a hand detection algorithms to segment adjacent object masks that are further used as training data for object detection. Further, if 3D data is available, ER based methods enable the the gesture aware \cite{Chidambaram.2024,Duver.17.01.202419.01.2024} or native controller based \cite{Franzluebbers.2022,FlorianWirth.2019} selection and segmentation of partial 3D point clouds to be used in ML training pipelines. In parallel, the annotation process has evolved into a significant business in its own right. The global data annotation market has experienced remarkable growth, with industry leaders such as Scale AI and Appen driving large-scale annotation operations. Annotation services are even expanding into low-income regions, with companies such as CloudFactory taking advantage of lower labor costs in emerging markets. Moreover, the effective use of advanced annotation tools typically requires a certain level of IT expertise \cite{lionbridge2021}, and the annotation process is often separate from the actual inspection activities, involving different workflows and personnel. Undoubtedly, the generation of training data represents a significant investment of both time and money, as highlighted by recent industry analyses \cite{gartner2022annotationcost}. 

\section{Concept Formulation} 

The hardware of our proposed system consists of a 5 MP industrial camera, a tracking system including a tracked pointer and a projector to provide visual feedback during the annotation process (Fig. \ref{fig:hardware_setup}). Both camera systems are mounted on a rigid support structure positioned 90 cm above the table surface with a 12 mm camera lens. While the projection can be considered an optional feature aimed at user convenience, early system tests and trials revealed that operators valued the visual feedback for a more confident and intuitive annotation workflow. To provide core functionality, we developed a software consisting of a simple front- and a back-end to handle the processing logic such as trajectory approximation and system communication. In principle, exporting annotation data is straightforward - one can choose a simple line-by-line format (e.g. normalised bounding box coordinates as used in YOLO), or opt for structured formats such as XML or JSON that embed detailed object attributes and spatial information. However, by integrating the CVAT API, we leverage its export options, editing capabilities and data consistency benefits. This integration streamlines our workflow and simplifies downstream processing, ensuring that the resulting annotations are usable for model training.\\
To derive image annotations such as polygons, polylines and bounding boxes from tracked 3D trajectories, we simply need to express single point measurements in pixel coordinates of a sufficiently high-resolution camera. Taking into account the proposed fully calibrated hardware setup (Fig. \ref{fig:hardware_setup}), this can be formulated as follows:
\begin{figure}[htbp]
\centering
\includegraphics[width=0.35\textwidth]{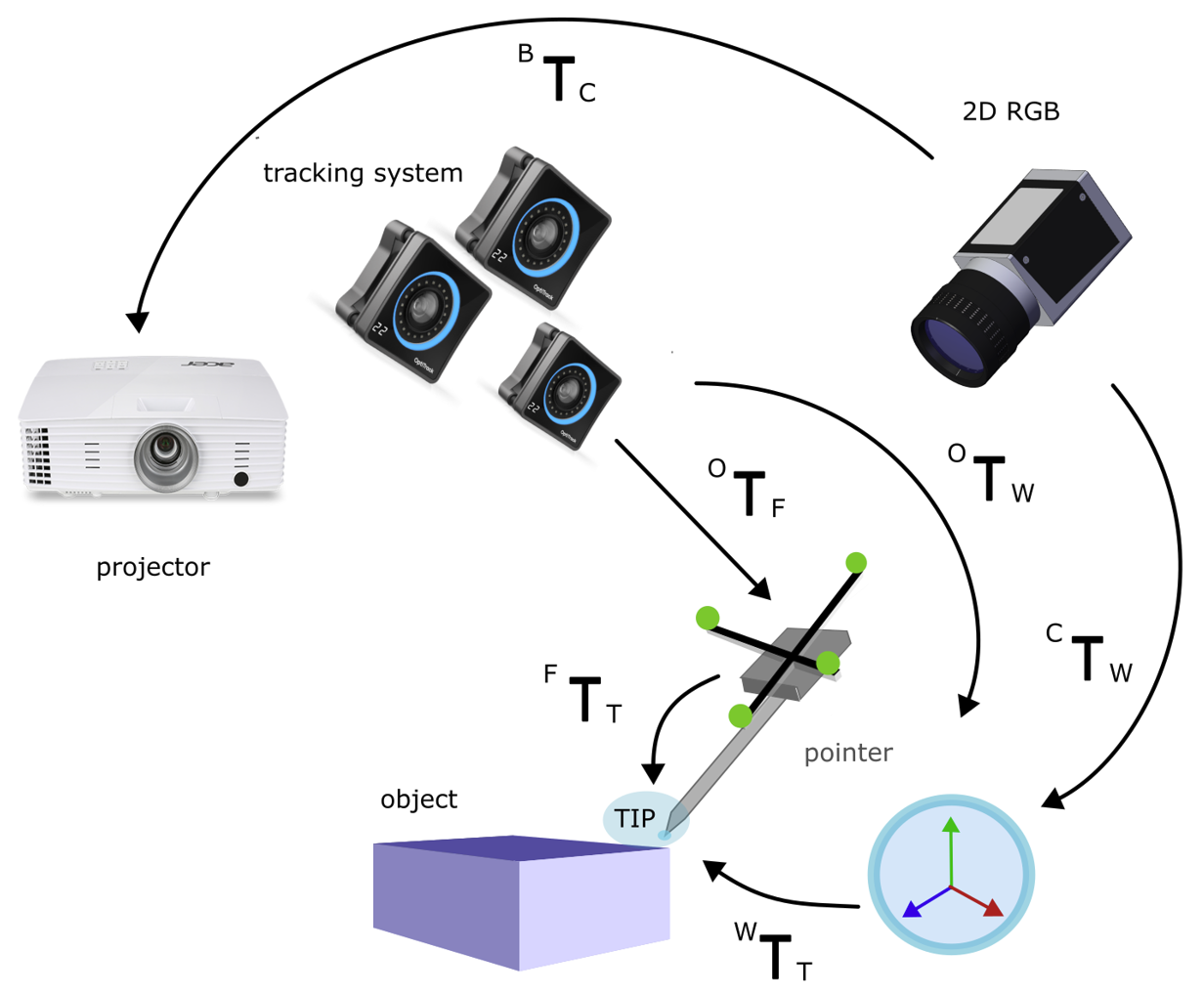}
\caption{Sketch of the hardware components used in the proposed system and all relevant transformations.}
\label{fig:hardware_setup}
\end{figure}

Let $\mathbf{p}_{\text{tip}}^P \in \mathbf{R}^3$ be the position of the tracked tip in the used coordinate frame of the pointer. The pivot calibration determines the fixed position of the tip in the pointer frame. Its expression in the world frame using homogeneous coordinates is given by

\[ \begin{pmatrix} \mathbf{p}_{\text{tip}}^C \\ 1 \end{pmatrix} = {}^C T_W{}^W T_P \begin{pmatrix} \mathbf{p}_{\text{tip}}^P \\ 1 \end{pmatrix}, \] where ${}^W T_P $ and ${}^C T_W $ are the calibrated transformations from pointer to world and from world to camera frame respectively, given by

\[ {}^W T_P = \begin{pmatrix} {}^W R_P & {}^W \mathbf{t}_P \\ \mathbf{0} & 1 \end{pmatrix} \text{ and } {}^C T_W = \begin{pmatrix} {}^C R_W & {}^C \mathbf{t}_W \\ \mathbf{0} & 1 \end{pmatrix}
\]
Let the tip in the camera frame be $\mathbf{p}_{\text{tip}}^C = \begin{pmatrix} X^C Y^C  Z^C \end{pmatrix}^T.$The pinhole camera model with camera intrinsics $K_C$ projects this point onto the image plane:
\[
s \begin{pmatrix} u_C \\ v_C \\ 1 \end{pmatrix} = K_C\begin{pmatrix} X^C \\ Y^C \\ Z^C \end{pmatrix} \text{ with } K_C = \begin{pmatrix}
f_x & 0   & c_x \\
0   & f_y & c_y \\
0   & 0   & 1
\end{pmatrix}.
\]
This yields the pixel coordinates:
\[
\begin{pmatrix} u_C \\ v_C \end{pmatrix} = \begin{pmatrix} f_x \dfrac{X^C}{Z^C} + c_x \\[1mm] f_y \dfrac{Y^C}{Z^C} + c_y \end{pmatrix}.
\]

To enable the projected visual feedback, the used projector with calibrated intrinsics $K_B$ is also extrinsically calibrated to the camera. Following the above mentioned transformation chain, we obtain a single equation that maps the tip position from the pointer frame to the projected pixel coordinates:
\[
s' \begin{pmatrix} u_B \\ v_B \\ 1 \end{pmatrix} = K_B \, {}^B T_C \, {}^C T_W \, {}^W T_P \begin{pmatrix} \mathbf{p}_{\text{tip}}^P \\ 1 \end{pmatrix}.
\]
This formulation enables the visualization of the pointer’s trajectory on the surface via the projector.

\begin{figure*}[htb]
\centering
  \includegraphics[width=0.935\textwidth]{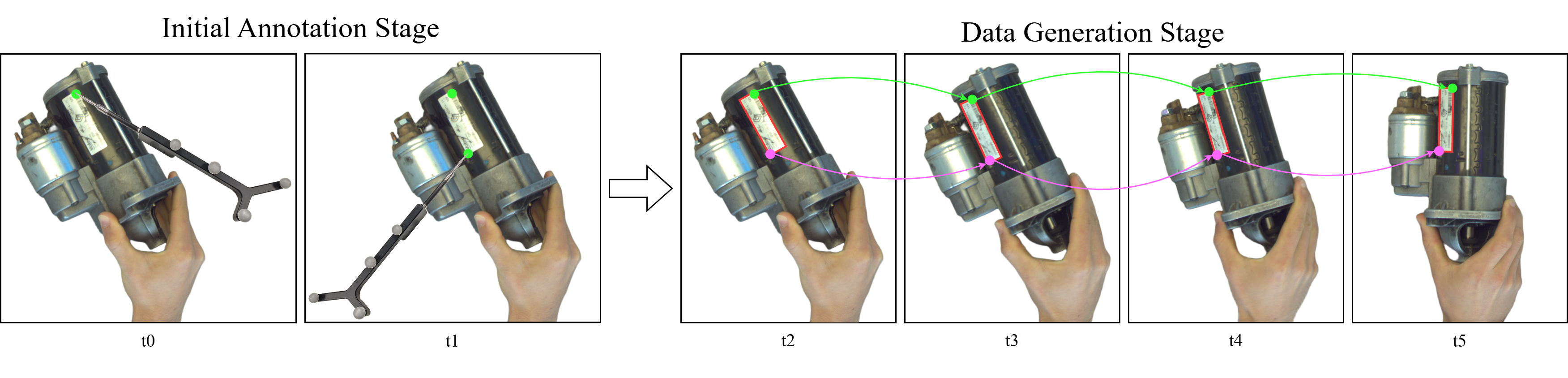}
  \caption{This diagram illustrates the two-stage process for training data generation. In this example, two points representing a rectangular surface area of the object are first selected to qualify the annotated area. Using optical flow to track both points, the desired data is then extracted from camera images when the pointer does not occlude the object.}
  \label{fig:multistage}
\end{figure*}
\section{System calibration}

Initially, a pivot calibration is performed to determine the centre of the pointer tip $c$ (a sphere with a diameter of $d=1$ mm) with respect to the dynamic reference frame (DRF) rigidly attached to the pointer device. The sphere centre in world coordinates $c'$ is described as: $\mathbf{c}' = \begin{bmatrix} x'_c,y'_c, z'_c \end{bmatrix}^T  = \mathbf{R}\,\mathbf{c} + \mathbf{t}$ where $\mathbf{R}$ and $\mathbf{t}$ represent the rotational and translation parts of the DRF's pose, respectively. $c$ can be calibrated by fixing the tip of the pen and collecting the DRF's pose at various positions $(\mathbf{R}_{k}, \mathbf{t}_{k})$. As the tip remains fixed, the problem can be formulated as: $\mathbf{R}_{k_1}\mathbf{c}+\mathbf{t}_{k_1} = \mathbf{R}_{k_2}\mathbf{c}+\mathbf{t}_{k_2}$. This forms a system of linear equations that can be solved to obtain the least squares estimate of the desired position $\mathbf{c}'$ \cite{Yaniv.2015}. To perform the transformations described in the previous section, the overall system (Fig. \ref{fig:hardware_setup}) must be calibrated. Besides the calibration of the camera's intrinsic $K_C$ \cite{888718}, the relative position of the camera with respect to the tracking system and the relative position of the projector with respect to the camera must be estimated. Both sensors are then expressed with respect to a fixed global coordinate system $W$ defined on the workspace surface. To calibrate the camera and tracking system, we use an asymmetric circle pattern. A small circular notch of 0.5 mm is precisely machined at the centre of each circle. By placing the pointer tip in the centre of each circle and recording its position as $\mathbf{c}'_n$, a set of 3D reference points is collected. Simultaneously, corresponding 2D image points are obtained from camera images using circle detection. This yields a Perspective-n-Point (PnP) problem that can be solved by minimising the reprojection error given by the sum of the squared deviations between the detected and reprojected circle centres \cite{LEVENBERG.1944}.
To fully calibrate the projector, we use the method proposed by Moreno et al.\cite{Moreno.2012}. Their method uses an extended pinhole model that assumes distortion for the projector. A printed checkerboard and a projected grey pattern sequence are used to obtain a dense set of point correspondences between the projector and the camera. The method introduces the concept of local homographies for a path around each checkerboard corner to calibrate the projector's intrinsic parameters. Finally, a stereo calibration is used to calculate projector's pose relative to the camera, denoted ${}^B\mathbf{T}_C$.

\section{Multistage Annotation Process} 

Due to the physical nature of the pointer device, occlusions inevitably occur when image data is recorded during trajectory generation (Fig. \ref{fig:multistage}). Consequently, this image data cannot be used for the annotations. To address this issue, we separate the annotation phase from the training data generation. Initially, the pointer is used on a stationary object to create trajectory data, which is then transferred to an image of the object at a time when the pointer was not visible, thus occluding the object. However, in this static case, only one training sample can be generated at a time. To create additional training samples, the object would need to be moved and the process repeated. As it is generally desirable to generate a large number of training samples from a single trajectory, we use an optical flow-based point-tracking approach to digitally 'glue' the projected annotation path onto the object's surface in every camera image taken during the data creation stage (Fig. \ref{fig:multistage}). Moving the object within the camera's field of view enables us to generate a large number of training samples from various angles. Moreover, extending to a multi-camera setup enables even more extensive data generation. A single, straightforward annotation can thus produce a rich dataset covering various perspectives, significantly enhancing the value of each annotation instance.

\begin{figure}[htbp]
\centering
\includegraphics[width=0.49\textwidth]{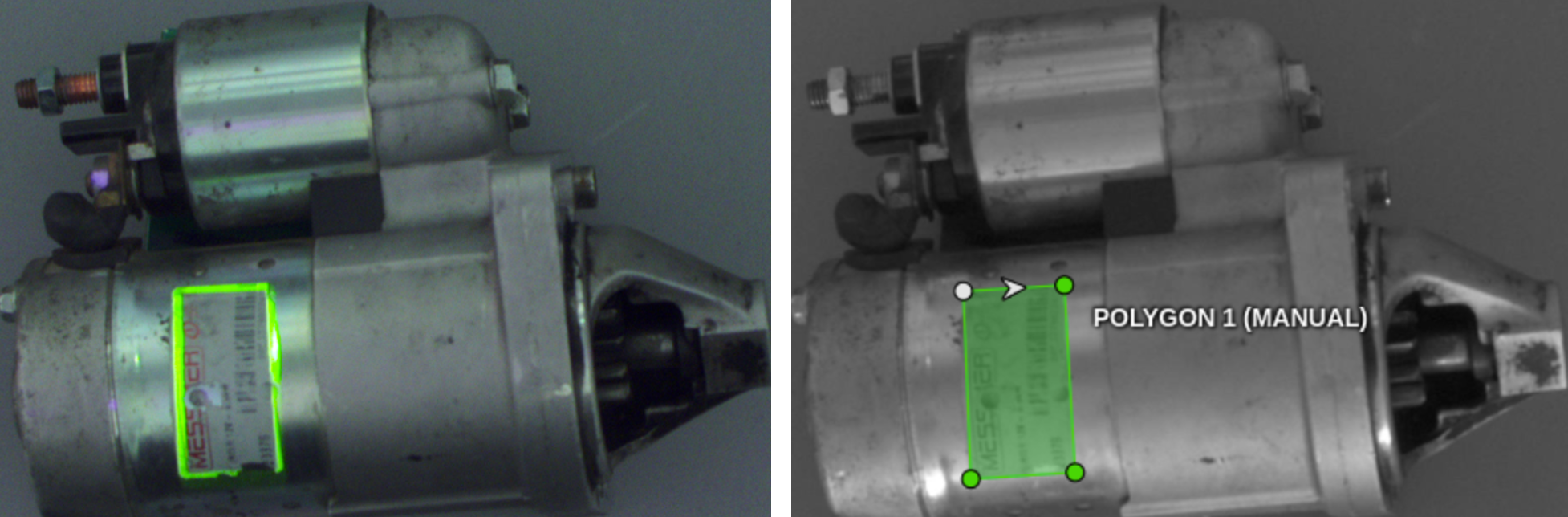}
\caption{A pointer-based created polygon projected onto the object (left). The resulting annotation for use in ML training pipelines visualized in CVAT \cite{cvat2020} (right).}
\label{fig:hardware_setup}
\end{figure}

\section{Discussion and Outlook}

This concept paper proposes a system that is capable of generating training data for specific areas of interest through direct physical interaction with the surface of an object that is subject to optical inspection.
Initial tests have demonstrated, that the introduced components and methods are entirely suitable for generating unrestricted training samples for ML pipelines in various scenarios. In the preliminary experiment, the tip of the pointer was navigated along a series of two-dimensional shapes on a flat surface, including a line, a hexagon, a rectangle, a sine curve, and two distinct irregular shapes. The deviation from the mean line was found to be 0.72 mm for regular shapes and 1.62 mm for irregular shapes, with these measurements obtained from 90 repetitions by three individuals. As the focus of this work lies primarily on the conceptual description of technical implementation details, the experimental evaluation of the tracking accuracy, annotation efficiency and the overall system usability are reserved for future research. These investigations will be essential for validating the system’s effectiveness and identifying areas for further optimisation in practical deployment.

\bibliographystyle{abbrv} 
\bibliography{literature} 

\end{document}